\documentclass{article}

\usepackage{arxiv}

\usepackage[utf8]{inputenc} 
\usepackage[T1]{fontenc}    
\usepackage{hyperref}       
\usepackage{url}            
\usepackage{booktabs}       
\usepackage{amsfonts}       
\usepackage{nicefrac}       
\usepackage{microtype}      
\usepackage{lipsum}

\usepackage{times}
\usepackage{latexsym}
\usepackage{graphicx}
\usepackage{amsmath}
\usepackage{multirow}
\usepackage{xcolor,colortbl}
\usepackage{caption}

\title{ETHOS: an Online Hate Speech Detection Dataset
\thanks{(co)winning CrowdFlower’s AI for Everyone Challenge for Q4 of 2017: https://prn.to/2KVWubz}
}

\author{
  Ioannis Mollas \\
  Aristotle University of Thessaloniki\\
  Thessaloniki, 54636, Greece\\
  \texttt{iamollas@csd.auth.gr} \\
   \And
  Zoe Chrysopoulou \\
  Aristotle University of Thessaloniki\\
  Thessaloniki, 54636, Greece\\
  \texttt{zoichrys@csd.auth.gr} \\
  \And
    Stamatis Karlos \\
  Aristotle University of Thessaloniki\\
  Thessaloniki, 54636, Greece\\
  \texttt{stkarlos@csd.auth.gr} \\
  \And
    Grigorios Tsoumakas \\
  Aristotle University of Thessaloniki\\
  Thessaloniki, 54636, Greece\\
  \texttt{greg@csd.auth.gr} \\
}

\begin{document}
\maketitle

\begin{abstract}
Online hate speech is a recent problem in our society that is rising at a steady pace by leveraging the vulnerabilities of the corresponding regimes that characterise most social media platforms. This phenomenon is primarily fostered by offensive comments, either during user interaction or in the form of a posted multimedia context. Nowadays, giant corporations own platforms where millions of users log in every day, and protection from exposure to similar phenomena appears to be necessary in order to comply with the corresponding legislation and maintain a high level of service quality. A robust and reliable system for detecting and preventing the uploading of relevant content will have a significant impact on our digitally interconnected society. Several aspects of our daily lives are undeniably linked to our social profiles, making us vulnerable to abusive behaviours. As a result, the lack of accurate hate speech detection mechanisms would severely degrade the overall user experience, although its erroneous operation would pose many ethical concerns. In this paper, we present `ETHOS’, a textual dataset with two variants: binary and multi-label, based on YouTube and Reddit comments validated using the Figure-Eight crowdsourcing platform. Furthermore, we present the annotation protocol used to create this dataset: an active sampling procedure for balancing our data in relation to the various aspects defined. Our key assumption is that, even gaining a small amount of labelled data from such a time-consuming process, we can guarantee hate speech occurrences in the examined material.
\end{abstract}


\section{Introduction}

Hate speech (HS) is a form of insulting public speech directed at specific individuals or groups of people on the basis of characteristics, such as race, religion, ethnic origin, national origin, sex, disability, sexual orientation, or gender identity\footnote{\url{https://en.wikipedia.org/wiki/Hate_speech}}. This phenomenon is manifested either verbally or physically (e.g., speech, text, gestures), promoting the emergence of racism and ethnocentrism. Because of the social costs arising out of HS, several countries consider it an illegal act, particularly when violence or hatred is encouraged~\cite{DBLP:conf/ijcai/DinakarPL15}. Although a fundamental human right, freedom of speech, it is in conflict with laws that protect people from HS. Therefore, almost every country has responded by drawing up corresponding legal frameworks, while research which is related to mechanisms that try to remedy such phenomena has recently been done by the Data Mining and Machine Learning (ML) research communities~\cite{JIROTKA2020100002}.

Another important issue is that the occurrence of HS phenomena is emerging in the social media ecosystem, distorting their initial ambition of favouring communication between their corresponding members independently of geographical restrictions and enriching similar activities~\cite{DBLP:conf/sai/TangC20}. The anonymity of social media is the primary explanation for the growth of such phenomena, as is the deliberate avoidance of subsequent legislation. Big companies, like Google and Facebook, are therefore obliged to remove such kind sof violent content from their platforms. As a result, Artificial Intelligence (AI) methodologies are employed to detect (semi-)automatically HS in real time, or even to prevent users from publishing similar content with appropriate warnings or bans. 

The solution of quarantining in an online fashion has recently been demonstrated~\cite{DBLP:journals/ethicsit/UllmannT20}, trying to smooth the censorship and the possible harmful consequences of HS attacks, while learning from short-text segments blooms in the last years~\cite{TOMMASEL20181}. Two of the most important features accompanying the short-text segments, sparseness and the presence of noise~\cite{DBLP:journals/ker/TommaselG19}, settle HS detection, a difficult task for the creation of fully automated solutions. Whereas problems of scalability arise when large quantities of data are simply collected without pre-processing or filtering. These points are of primary importance to this work.

To achieve high performance in real-world tasks, AI methodologies require balanced, accurate, and unbiased datasets. This requirement, however, is rarely met without applying proper annotation stages~\cite{DBLP:conf/cikm/ChenMLZM19,ibrohim-budi-2019-multi}. This is the direction in which our work aims to make a significant contribution, motivated by the HS use case, providing also a generic-based protocol that could be extended to a wide variety of learning tasks. To be more precise, the relevant literature currently contains a large number of manually created HS datasets~\cite{waseem-hovy,DBLP:conf/naacl/ZampieriMNRFK19}. However, since the majority of them were not carefully collected during the corresponding sampling stages, they are essentially large sets of annotated samples on which undesirable phenomena occur frequently. Specifically, highly imbalanced classes or redundant information prevent the subsequent implemented learning models from effectively harnessing the underlying patterns. 

\begin{figure}[!ht]
    \centering
    \includegraphics[width=0.85\textwidth]{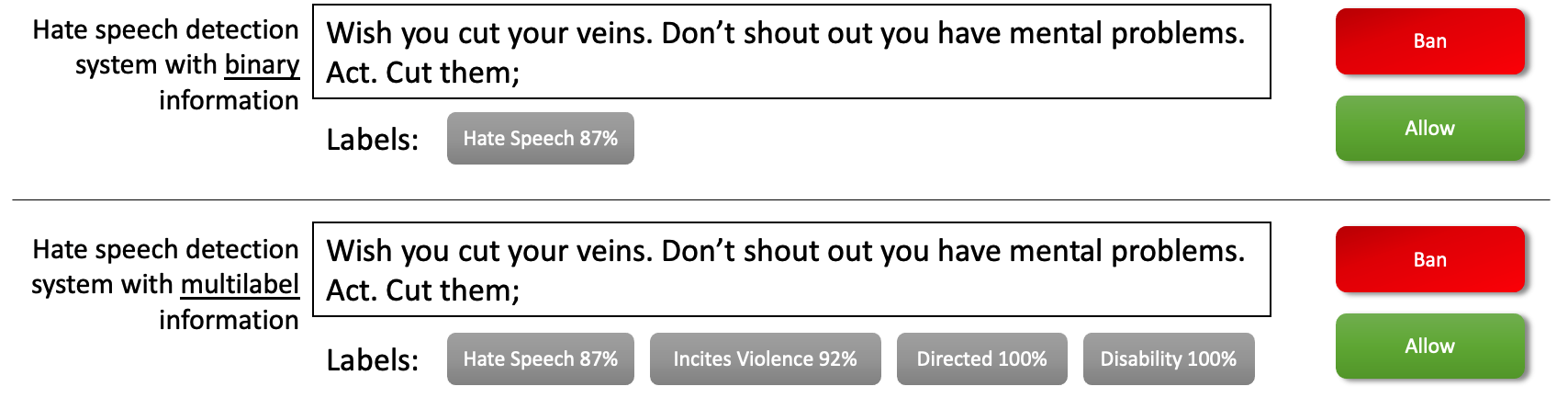}
    \caption{A realistic example of informing a human reviewer about an investigated comment on binary (top) and multi-label (bottom) level}
    \label{fig:labels}
\end{figure}

Moreover, by sampling the regions of feature space which express only a restricted level of uncertainty when unlabelled data are queried may settle the learning strategy myopic. All these phenomena violate the previously specified desired requirements resulting in solutions with low variance and/or high bias~\cite{DBLP:conf/lrec/RosendaalCN20}. Furthermore, most of them are concerned with binary or multi-class classification tasks, while overlooking the more practical case of multi-label classification (MLL). Label dependencies and the semantic overlap that occurs on MLL cannot be ignored when protection from hateful comments is the main task. Since an online comment can fit to more than one defined label at the same time, rather than being limited to just one outcome, investigation of the latter scenario appears to be more effective (see Figure~\ref{fig:labels}). This aspect is also studied here because the difficulties described previously are enforced under the MLL scenario.

A simple application that uses the MLL schema provided by the proposed HS dataset could be an assistance system for human staff reviewing comments on social media platforms. This would make it easier for the reviewer(s) to decide if the message contains HS content by providing more insights. For example, if a comment is presented as targeting people with disabilities, directed at a person, and encourages violence, it will be more helpful for the reader to conclude and condemn it for containing HS rather than being presented with a single label (e.g., `may contain HS':\{`yes',`no'\}). In terms of the ethical issues that emerge in the case of HS, it appears that a proper manipulation protocol is required for preventing possible defects. Such protocols have addressed wider or more focused research topics, such as news articles, although similar directions have recently been explored in the field of HS detection~\cite{DBLP:journals/corr/abs-2004-14454}. 


In this paper, we present the process of creating a multi-labelled dataset with a step-by-step narrative, to avoid the consequences that typically occur in attempts with data that depend on social media platforms, and to increase the likelihood of mining more informative instances. Although the design of the proposed protocol can fit with any target domain indisputably, we are currently focusing on addressing the HS scenario and provide some insightful analysis of this use case. In this attempt, an existing dataset mined from popular social media platforms has been exploited, while a well-known crowdsourcing platform was used for validating the final result. The proposed annotation protocol's effects are discussed in detail and visualised using explanatory methods. Following that, a series of experiments are being conducted to determine the baseline performance of this particular dataset using state-of-the-art (SOTA) techniques. From traditional ML algorithms and ensemble models to neural networks (NNs) with and without embeddings (\emph{emb}) information, binary and multi-label experiments have been performed, inspired primarily by other similar approaches to presenting research datasets~\cite{almeida2013towards,hateTweets,ljubesic-etal-2018-datasets}. Despite the limited size of the investigated dataset, its careful design during the active sampling stage and the consistency of he included samples were proven beneficial based on our results. 

Our ultimate ambition, by describing the total procedure and providing the corresponding dataset, is to foster any interested researchers/businesses to take into consideration an approach that attempts to transform the existing insulting environment of social media into a non-hate, inclusive online society. Adoption of the proposed annotation protocol into different scientific fields could prove quite beneficial, especially when the knowledge acquired by oracles during annotation may be ambiguous. The assets also gained by examining the HS problem through a multi-label view help us clarify the harasser's actual motivations and lead to more targeted comments when dedicated platforms try to inform the corresponding victims~\cite{DBLP:conf/ijcai/DinakarPL15}. And, of course, the insights gained through such protocols could enhance the ability of ML learners to generalise when applied to different datasets that contain similar classification categories, despite the limited size of the proposed dataset over which they are trained. The proposed strategy of actively creating a balanced dataset, preserving the informativeness of each class and minimising the redundancy of the included instances, constitutes the key asset of our protocol. Our in-depth experiments support our hypotheses, particularly regarding the most difficult classes to detect.

The rest of this paper is structured as follows: Section 2 includes several well-documented attempts to address the HS problem using samples gathered from related sources. The proposed annotation protocol is defined next, followed by some extended single/multi-label classification experiments in Section 4, which demonstrate the discriminating ability of several algorithms under consideration. Section 5 presents a few studies with a variation of the original dataset and two additional datasets. Finally, Section 6 discusses the more crucial assets of the proposed dataset, and the annotation protocol, also regarding the recorded experiments, reporting later some remarkable future points that could be further investigated.

\section{Related Datasets} 
\label{sec:relDat}
In this section, we present datasets related to HS, along with their formulation as well as some useful information about their structure and/or the manner under their composition took place. The last paragraph describes the Hatebusters' data that we utilise as a seed data through the proposed protocol to produce the final structure of data, named {\em ETHOS} (onlinE haTe speecH detectiOn dataSet).

A collection of 16.914 hate speech tweets was introduced in a study of how different features improve the identification of users that use analogous language online~\cite{waseem-hovy}. Out of the total number of messages, 3.383, 1.972 and 11.559 concerned sexism, racism and did not include HS, respectively, while were sent by 613, 9 and 614 users. The corpus was generated by a manual tweet search, containing popular slurs and terms related to sexual, religious, gender and ethnic minorities in order to include samples that are not offensive regardless of the inclusion of such words. The main drawback here is the access to the text of the tweets only through the public Twitter API.

Another dataset (D1)~\cite{hateTweets} contains 24.783 tweets, manually classified as HS (1.430), offensive but not HS (19.190), and neither hate nor offensive speech (4.163) by Figure-Eight's\footnote{Formerly Crowdflower and latterly Appen: \url{https://appen.com/figure-eight-is-now-appen/}} members. The data was gathered again via the Twitter API, filtering tweets containing HS words submitted to Hatebase.org. The outcome was a sample of 33.548 instances, while 85.4 million tweets were collected from the accounts of all users. A random sample of this collection led to the final dataset. Nevertheless, this dataset lacks diversity in terms of HS content. For example, the gender-based HS tweets are biased towards women, while the greatest number of them contain ethnicity content. 

Research focusing on the identification of misogynistic language on Twitter uses a dataset called Automatic Misogyny Identification (AMI)~\cite{fersini2018overview} with 4.000 annotated comments and binary labels. Apart from this labelling mode, every comment is defined by two extra fields. The first one concerns the type of misogynistic behaviour: stereotype, dominance, derailing, sexual harassment, discredit or none (if the tweet is not misogynous). The second one concerns the subject of the misogynistic tweet: active, when it attacks a specific target (individual), passive, when it denotes potential receivers (generic), and again none, if there is no misogyny in the tweet.

The largest online community of white nationalists, called Stormfront, was used to form another dataset~\cite{whitesupremacist}. The content in this forum revolves around discussions of race, with various degrees of offensiveness, included. The annotation of the samples is at the sentence level, which is a technique that keeps the smallest unit containing hate speech and reduces noise. The dataset contains 10.568 sentences that are classified as HS (1.119 comments) or not (8.537 comments), as well as two supplementary classes, {\em relation} for sentences that express HS only when related to each other and {\em skip} for sentences which are not in English or do not contain any information as to be accordingly classified. Furthermore, information like the post identifier and the sentence’s position in the post, a user identifier and a sub-forum identifier, as well as the number of previous posts the annotator had to read before making a decision over the sentence’s category are also recorded. The samples were picked randomly from 22 sub-forums covering diverse topics and nationalities. 

A dataset introduced by Fox News~\cite{foxnews} consists of 1.528 Fox News users' comments (435 hateful), which were acquired from 10 discussion threads of 10 widely read Fox News articles published during August 2016. Context information is considered extremely important, so details such as the screen name of the user, all the comments in the same thread and the original article, are also included.


A recent multilingual work (D2)~\cite{DBLP:conf/emnlp/OusidhoumLZSY19}, a trilingual (English, French and Arabic) dataset with tweets, was created attempting to mine similar expressions of 15 common phrases over these languages, focused on different sources of obscene phrases (e.g. more sensitive topic-based discussions based on locality criteria). After tackling some linguistic challenges per separate language, and a strict rule set that was posed to human annotators from the Amazon Mechanical Turk platform to ensure trustworthy feedback, a pilot test set was provided. Having gathered the necessary evaluations, another one reconstruction of the label set was applied, before the final formulation of 5.647 English, 4.014 French and 3.353 Arabic tweets was reached, annotated over 5 separate tasks. Apart from the binary directness of each tweet that was tackled better by single task language models, the rest 4 classification tasks, which included at least 5 label gradations, were clearly boosted via multi task single/multi language or single/multi multilingual models.

The issue of cyberbullying has been recently investigated also, where the skewed distribution of positive and negative comments was tackled by tuning a cost-sensitive linear SVM learner over various combinations of joined feature spaces, and obtaining similar performance on both English and Dutch corpus~\cite{VanHee2018}. An investigation of recognising also the role of each participant during such phenomena took place, while a qualitative analysis raised the difficulty of reducing misclassification scores when irony exists in offensive comments.

Finally, a small collection of 454 YouTube comments annotated as HS (120) or not (334) was introduced by the creators of the Hatebusters Platform~\cite{hatebusters}, which aims to establish an online inclusive community of volunteers actively reporting illegal HS content on YouTube. This dataset, through semi-supervised learning, was evolving in the Hatebusters Platform improving the predictivity of the ML models. However, this unpremeditated expansion of the dataset led to a more redundant variant of its original form. We use the initial collection of Hatebusters' data as a seed to the protocol that we propose in the following section.


\begin{figure}[!ht]
    \centering
    \includegraphics[width=0.85\textwidth]{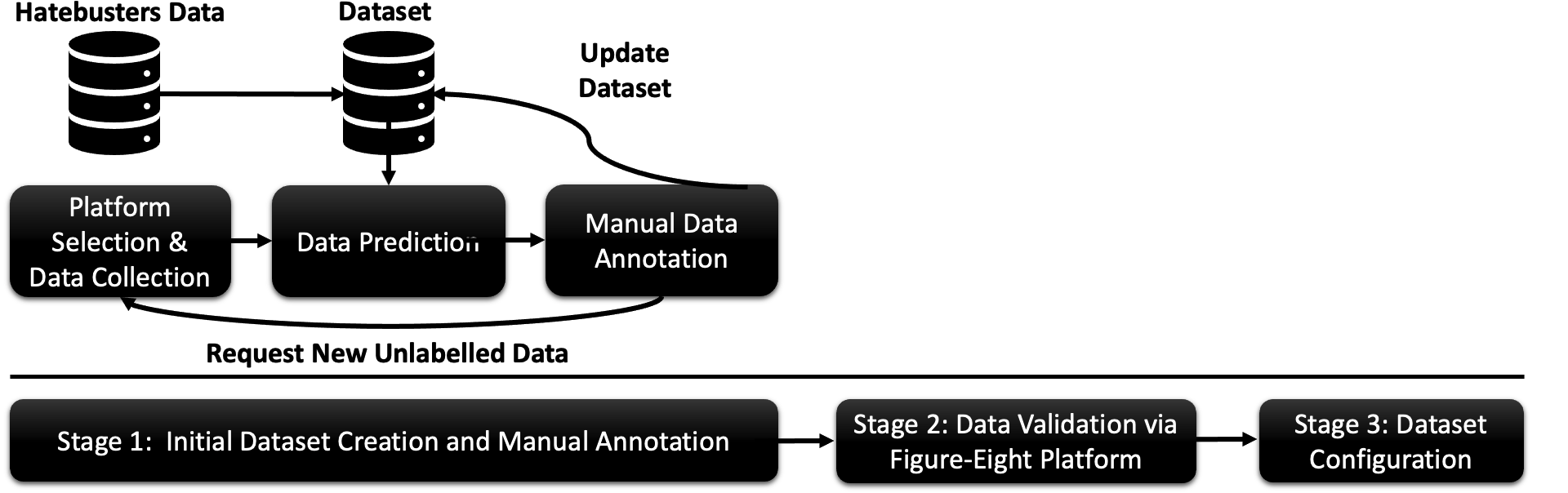}
    \caption{Dataset creation stages flowchart}
    \label{fig:dcreation}
\end{figure}

\section{ETHOS Dataset Creation}
To overcome the key weaknesses of the existing collections of HS instances, we introduce a small, yet fairly, informative dataset, ETHOS, that does not suffer from issues such as imbalanced or biased labels (e.g., gender), produced appropriately following the proposed protocol. Considering the aforementioned popular approaches of mining similar datasets for tackling with HS problem, we assume that an appropriate pre-process of initially collected data could improve in general their overall utilisation under ML or AI products, improving the total fitness of data quality, blending data mining techniques related with the field of Active Learning~\cite{DBLP:conf/naacl/SharmaZB15}, such as query strategy and crowdsourcing platforms. The overview of the proposed annotation protocol is visualised through a flow chart in Figure~\ref{fig:dcreation}. The finally obtained dataset is the outcome of a 3-stage process, which we describe shortly in the current Section.

\subsection{Initial Dataset Creation and Manual Annotation}
The first three procedures, mentioned as ``Platform Selection \& Data Collection'', ``Data Prediction'' and ``Manual Data Annotation'', could be seen as the initial stage (Stage 1) which is executed until a stopping criterion is satisfied regarding the cardinality of the collected instances, based on the original available HS dataset which operates as the input. This stage works like a ``stream'', specifically for groups of comments that we have already collected, annotating their weak labels' predictions through a predefined ML classifier, before an active selection and manually annotation takes place over some unlabelled ($U$) mined examples.

\subsubsection{Platform Selection \& Data Collection}
To create this dataset ($D$), initially $D = \emptyset$, a 
data collection protocol has been designed. We chose the platforms of Hatebusters\footnote{\url{https://hatebusters.org}} and Reddit through the Public Reddit Data Repository\footnote{\url{https://files.pushshift.io/reddit/comments/}} to collect our data. Hatebusters Platform collects new data daily via the YouTube Data v3 API
. 
After these new data have been collected, the Hatebusters Platform performs the classification process. The locally retained pre-trained ML model predicts the class of each comment, exporting a `hate' score. Currently, this model is a Support Vector Machine (SVM)~\cite{SVM} model with a linear kernel embedded with the well-known vectorization technique of the term frequency-inverse document frequency (TF-IDF). Instead of transforming the output of the SVM learner to confidence score, we kept its inherent property to compute the distance from the decision boundary. Through this, lower time overheads and more faithful decisions are drawn.

After granting access to Hatebusters' SQL database, based on the input data, this first part was to query the Hatebusters' database for comments already annotated by the corresponding users, without spending any monetisation resources. These comments were deemed to be accurate, and they were the first group of comments to be manually annotated. The second part concerns the enrichment of the gathered comments, by querying Hatebusters' database with a specific frequency (e.g. daily) for a time period -- in our case this was equal to two months -- with various queries. Based on the data obtained each previous day, the applied query strategy had been updated concerning only them. For example, when we received a sufficient amount for all categories of HS, except for one category, the queries in the Hatebusters' database were updated to make comments specific to the residual category. Later on, we will show the categories and the amount of comments we have received.

The final part of the 
data collection process was based on a public Reddit data archive, which provides batches of files regarding Reddit comments on a monthly basis. The files of this directory were processed through a JSON crawler for selecting comments from specific subreddits for particular time periods. The discovery of subreddits incorporating different HS contents has been investigated\footnote{\url{https://en.wikipedia.org/wiki/R/The_Donald}}$^,$\footnote{\url{https://en.wikipedia.org/wiki/Incel}}, 
we distinguished the next entities:
\begin{itemize}
    \item \textbf{Incels}, this subreddit became known as a place where men blamed women for their unintended celibacy, often promoting rape or other abuse. Those posts had a misogynistic and sometimes racist content.
    \item \textbf{TheRedPill}, which is devoted to the rights of men, containing misogynous material.
    \item \textbf{The\_Donald}, a subreddit where the participants create discussions and memes supportive of U.S. President Donald Trump. This channel has been described as hosting conspiracy theories and racist, misogynous, Islamophobic, and antisemitic content.
    \item \textbf{RoastMe}, in this subreddit, reddit users can ask their followers to `roast' (insult) them.
\end{itemize}

While some of these subreddits were suspended and shut down by Reddit at the end of 2017 due to their context, it was possible to access comments from these subreddits by selecting files from the archive for October 2017 and earlier.

\subsubsection{Data Prediction}
The next process of Stage 1 is the ``Data Prediction''. For each batch of comments extracted from the first part, the assignment of some useful labels to the available unlabelled set ($U^{current}$) is triggered through an ML model trained on an expanded version ($L \cup D$) of the Hatebusters' dataset ($L$) and the new data annotated on Stage 3 ($D$). 
Per each iteration of the previous part, we were performing a grid search among a bunch of classification methods in the currently expanded dataset, obtaining the best algorithm through a typical 10-fold-CV process so as to be set as the annotator of the ($U^{current}$).

The selected bunch consisted of various ML models: SVMs, Random Forests (RF)
, Logistic Regression (LR)
, as well as simple or more complex architectures of Neural Networks (NNs). In addition to the classifier tuning, some TF-IDF vectorization techniques -- with word or char $n$-grams ($n$ from 1 to 13) -- were also examined in this search.

\subsubsection{Manual Data Annotation} 
By the end of the ``Data Prediction'' phase, the ``Data Annotation'' process is initiated. In the sense of active learning concept, a hybrid combination of query strategy has been employed in order to pick informative comments for manual annotation. The mentioned query strategy combines appropriately both concepts of Uncertainty Sampling and Maximum Relevance with predefined ranges of accepted confidence values based on the expected labels of the classifier we had trained~\cite{DBLP:journals/kbs/PupoAV18}. More specifically, we were annotating the comments within the $[.4, .6]$ probability range, while we were examining few comments in the ranges $[.0, .1] \cup [.9, 1.0]$ to detect any major misclassification. Eventually, only comments with specific labels and content were added to the new dataset ($D$), preserving both the \textit{balance of the labels} and the \textit{diversity of the comments per label}. 

At the end of this process, if the number of comments collected is not more than a targeted threshold ($T$) -- in our case $T = 1.000$ -- we update the $D$, and Stage 1 will be repeated to request new unlabelled comments. Otherwise, Stage 2 will be triggered. Despite the limited cardinality of the exported dataset, the adopted actively sampling process eliminates defects of redundancy, maintaining the both informativeness of each label, and reducing at the same time overfitting phenomena. The issue of obtaining a myopic strategy is also eliminated, since different regions of uncertainty are explored~\cite{DBLP:journals/ml/KremplKL15}. The efficacy of such methods has been highly declared in the literature ~\cite{DBLP:journals/puc/MalikS17,DBLP:journals/jcst/KumarG20}. Therefore, an in-depth evaluation stage regarding several learning models has been conducted in Section 4.

\subsection{Data Validation via Figure-Eight Platform}
The second stage will begin when $T$ -- in our case $1.000$ -- comments have been collected. Moreover, Hatebusters' dataset is discarded, since it does not further contribute to our protocol. After a number of different experiments on the Figure-Eight platform, we settled on the next process. Firstly, given a specific comment, we ask the contributors to identify whether that comment \textit{contains HS or not}. In a positive scenario, we raise 3 more questions: whether the comment \textit{incites violence}, defining violence as ``the use of physical force to injure, abuse, damage, or destroy'', and whether the comment includes \textit{directed} or \textit{generalized} HS. The case of targeting a single person or a small group of people is defined as directed HS, whereas the case of targeting a class/large group of people is described as generalised HS. Finally, we ask the contributors to pick \textit{one} or \textit{more} from the following \textit{HS categories}, which, according to their opinion, better reflect(s) the content of the comments. The categories of HS concern gender, race, national origin, disability, religion and sexual orientation.

\begin{figure}[ht]
\centering
\includegraphics[width=0.45\textwidth]{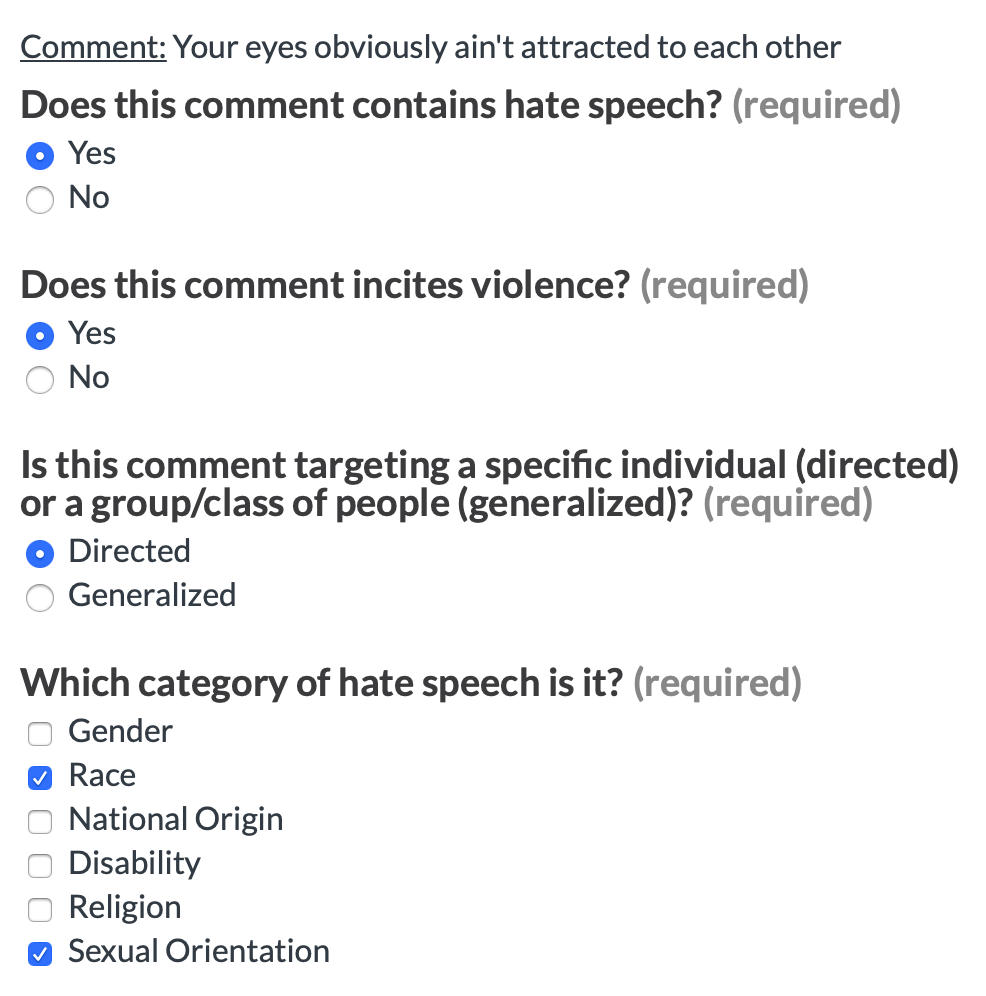}
\caption{Example of Figure-Eight's platform design}
\label{fig:figureEightExample}
\end{figure}

Through Figure~\ref{fig:figureEightExample}, an example of the designed query template on the crowdsourcing platform is presented, depicting the types of questions and potential responses. With respect to the proposed template, we introduced a set of 40 questions for validating the trustworthiness of human annotators, as well as a test run of 100 unlabelled comments, with 5 different contributors determining the labels for each one of these. The annotators were frequently reminded that there was the possibility of explicit material in the comments through the crowdsourcing platform. The feedback we collected helped us to finalise some design and instruction issues. Most importantly, we have fixed some naive and few serious mistakes on the validation questions, such as the following:

\textit{[naive mistake]}: For the comment ``J**s are lower class pigs.'' we had `Yes' on question `Does this comment incite violence?', while we gave as an explanation ``It spreads hatred, but not violence''.

\textit{[serious mistake]}: We misclassified the comment ``Q****s are an abomination and need to be helped to go straight to H**l!'' to the wrong category. By receiving the feedback we fixed it.

Then, we executed the task for the whole $D$, collecting in total 5.360 judgements. Almost every comment was therefore annotated by five different annotators. The level of expertise of the annotators was the 3rd, on a scale of 3 levels. ``The 3rd level annotators are the smallest group of the most experienced, most accurate, contributors'' according to the Figure-Eight System. We also computed the Fleiss' kappa, a statistical measure for assessing the reliability of agreement of annotators, and we present the results in Table~\ref{tab:aggrement}. A kappa value greater than 0.75 implies good agreement, while kappa values greater than 0.90 indicate perfect agreement~\cite{minitab}.

\begin{table}[ht]
\centering
\resizebox{0.9\textwidth}{!}{%
\begin{tabular}{c|c|c|c|c|c|c|c|c|c|}
\cline{2-10}
 & \begin{tabular}[c]{@{}c@{}}Contains \\ Hate Speech\end{tabular} & Violence & \begin{tabular}[c]{@{}c@{}}Directed vs \\ Generalized\end{tabular} & Gender & Race & \begin{tabular}[c]{@{}c@{}}National\\ Origin\end{tabular} & Disability & \begin{tabular}[c]{@{}c@{}}Sexual\\ Orientation\end{tabular} & Religion \\ \hline
\multicolumn{1}{|c|}{\begin{tabular}[c]{@{}c@{}}Fleiss'\\ Kappa\end{tabular}} & 0.814 & 0.865 & 0.854 & 0.904 & 0.931 & 0.917 & 0.977 & 0.954 & 0.963 \\ \hline
\end{tabular}%
}
\caption{Reliability of annotators agreement per label}
\label{tab:aggrement}
\end{table}

\subsection{Dataset Configuration}
The final stage regards dataset configuration. Taking as input the results from the Stage 2, the dataset takes its final form. Examining the annotated data one last time manually, we checked for any misclassification. Few errors occurred on some of the most disambiguous examples, assuring us about the quality of the annotators that participated. Although the Figure-Eight platform provides several attributes for informing suitably the human annotators, even stricter measures should be taken into consideration when large-scale datasets are aimed to be obtained~\cite{DBLP:conf/lrec/HaagsmaBN20}. 

The use of representative test questions that follow a more realistic label distribution than the uniform could be useful to the overall process. This might be improved further by incorporating an interactive procedure that alerts annotators to mislabelled samples and/or allows them to provide feedback when they disagree. Despite the inherent uncertainties introduced by the human factor, crowdsourcing is the sole viable technique for gathering the required information regarding the label space. This is true not only for large-scale datasets, but also for smaller cases~\cite{DBLP:conf/eacl/NghiemBA21}. 

Furthermore, given the semantic overlap of label space encountered during HS detection, the assumption of obtaining cheap labels is violated. Given the idiomatic expressions and highly unstructured nature of the comments posted on social media platforms, this becomes especially clear when examined in a multi-label fashion. To address this, additional human supervision, as stated at this stage, is required, while the active sampling process, which aims to create a balanced dataset, is clearly justified.

\subsection{ETHOS Dataset Overview}

Two datasets\footnote{\url{https://github.com/intelligence-csd-auth-gr/Ethos-Hate-Speech-Dataset.git}}
were the product of the above operation. ``Ethos\_Binary.csv'', the first one, includes 998 comments and a label on the presence or absence of hate speech content (`\textit{isHate}'). The second file, called ``Ethos\_Multi\_Label.csv'', includes 433 hate speech messages along with the following 8 labels: (`\textit{violence}', `\textit{directed\_vs \_generalized}', `\textit{gender}', `\textit{race}', `\textit{national\_origin}', `\textit{disability}', `\textit{sexual\_orientation}', `\textit{religion}'). 

For every comment $c_i$, $N_i$ annotators voted for the labels that we set. The label `\textit{isHate}' was the result of summing up the positive votes $P_{1,i}$ of the contributors, divided by $N_i$, so its values are within the range of $[0,1]$. We measured the `\textit{violence}' label by summarising the positive votes of the contributors $P_{2,i}$ to the question: ``Does this comment incite violence?'', which was divided by $P_{1,i}$ to be normalised to $[0,1]$. Likewise, the value of the label `\textit{directed\_vs\_generalized}' was determined by summarising the annotators replied `directed' $P_{3,i}$ to the question, ``Is this comment targeting a specific individual (directed) or a group/class of people (generalized)?'', divided by $P_{1,i}$. Finally, we accumulated the votes of the $P_{1,i}$ contributors for each of the 6 hate speech categories, and dividing them by $P_{1,i}$, we obtained six independent labels.

\begin{figure}[ht]
\centering
\minipage{0.4\textwidth}
\centering
\includegraphics[width=0.8\textwidth]{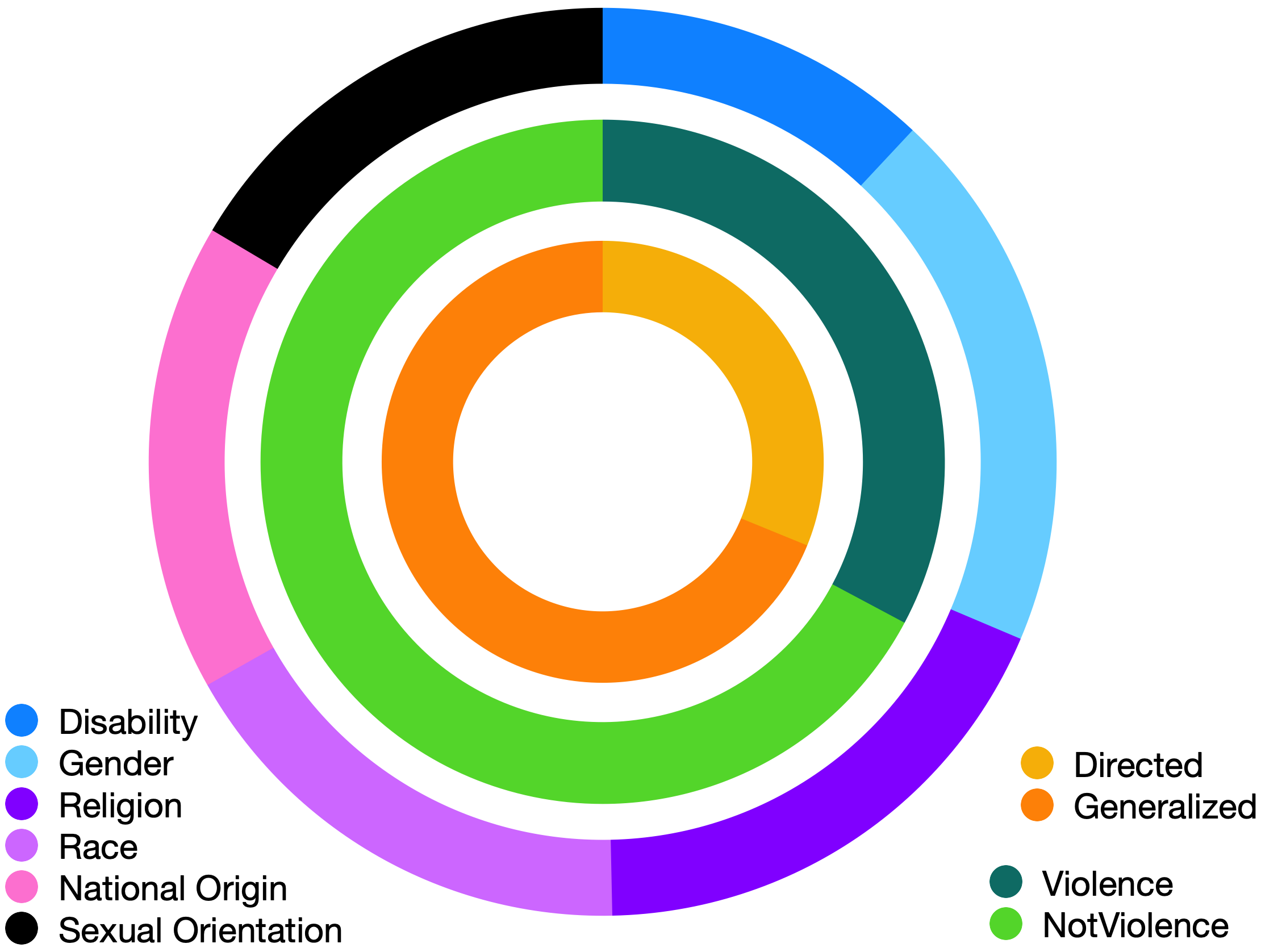}
\caption{Ratio of labels}
\label{fig:isHateLabels}
\endminipage\hfill
\minipage{0.57\textwidth}
\centering
\resizebox{0.8\textwidth}{!}{%
\begin{tabular}{c|c|c|c|c|c}
\cline{2-5}
 & V-D & nV-D & V-G & nV-G & \\ \hline
\multicolumn{1}{|c|}{Gender} & 14 & 22 & 13 & 37 & \multicolumn{1}{c|}{86} \\ \hline
\multicolumn{1}{|c|}{Race} & 4 & 13 & 12 & 47 & \multicolumn{1}{c|}{76} \\ \hline
\multicolumn{1}{|c|}{National Origin} & 5 & 11 & 18 & 40 & \multicolumn{1}{c|}{74} \\ \hline
\multicolumn{1}{|c|}{Disability} & 12 & 15 & 8 & 18 & \multicolumn{1}{c|}{53} \\ \hline
\multicolumn{1}{|c|}{Religion} & 11 & 8 & 24 & 38 & \multicolumn{1}{c|}{81} \\ \hline
\multicolumn{1}{|c|}{Sexual Orientation} & 11 & 15 & 11 & 36 & \multicolumn{1}{c|}{73} \\ \hline
 & 57 & 84 & 86 & 216 & \multicolumn{1}{c|}{443} \\ \cline{2-6}
\end{tabular}%
}
\captionof{table}{Correlation of HS categories with (not) violence (nV - V) and directed/generalized (D - G) labels}
\label{tab:correlTable}
\endminipage
\end{figure}

This dataset achieves to create balanced labels. In particular, it maintains balance between the two classes of `isHate' label, almost perfect balance between the 6 labels of hate speech categories, while it has a fair ratio between the rest of the labels (Figure~\ref{fig:isHateLabels}). In Table~\ref{tab:correlTable}, the balance between hate speech categories (last column) and their correlation with violence and directed/generalized labels is further portrayed.

\section{Dataset Baseline Evaluation}
In order to evaluate ETHOS, after pre-processing the data, we used a variety of algorithms in binary/multi-label scope to present the baseline performance in this dataset. For the purpose of providing the unbiased performance of each algorithm we performed nested-CV~\cite{nestedCrossValidation} evaluation, using a variety of parameter setups, for every algorithm except NNs, where we applied 10-fold-CV~\cite{crossValidation}. In addition, we binarise the values of each label, which are initially discrete in a range of [0,1], to the \{0,1\} classes using the rule \textit{``If value $\geq 0.5 \to 1$ Else value $\to 0$''}. More in-depth details follow next.

\subsection{Data Preparation}

The pre-processing methodology used in our case begins with lowercasing transformation, contraction transformations (available into the zip file), removal of punctuation marks, and stemming and lemmatization via Snow-ball stemmer~\cite{snowball} and WordNet lemmatizer~\cite{wordnet}.

Before we proceed to the experiments, we transform the pre-processed textual data into word vectors using TF-IDF and Text-to-Sequences processes. Particularly, for the former, several parameter tuples of (n\_gram, max \_features, stopwords existence) were examined, while on the latter, the corresponding number of maximum features was set at 50k. Moreover, 3 pre-trained models that concern computation of \emph{emb} were included: FastText (FT)~\cite{fastText}, GloVe (GV)~\cite{glove}, Bert Language Model (BERT)~\cite{bert}, and the distilled version of BERT (DistilBERT)~\cite{distilBert}. We should mention that the steps of stemming and lemmatization were skipped in the Text-to-Sequence experiments.

\subsection{Binary Classification}
A lot of applications are investigating the problem of HS detection through a binary scope. It is therefore necessary to present the performance of SOTA algorithms on such a version of this dataset.

Thus, we used the following algorithms for our experiments in this stage: Multinomial and Bernoulli variations of Naive Bayes (MNB and BNB, respectively)~\cite{NB2}, LR, SVMs, RF and Gradient Boosting (Grad)~\cite{GBoost}. Moreover, we used four different NN architectures, as other similar works attempt~\cite{DBLP:journals/corr/abs-2003-07459}. The first one utilises convolutional NNs (CNNs)~\cite{CNNnets} with an attention~\cite{attentionLayers} layer. A single LSTM-based NN constitutes the second architecture. The third model is an NN with multiple parallel layers, which contain CNNs, LSTMs and FeedForward layers (FFs). The last architecture consists of Bidirectional LSTMs (BiLSTMs). We combined these NNs with FT and GV. Lastly, we used BERT and DistilBERT, which were fine-tuned in our classification task. Such architectures have met great acceptance in the related ML community~\cite{DBLP:conf/clic-it/PolignanoBGSB19,yang-etal-2019-exploring-deep}.

We chose accuracy and precision, recall and $F_1$-score with macro indication, and the confusion matrix as metrics.Furthermore, we calculate specificity $TN/N$ and sensitivity $TP/P$. However, in applications like HS monitoring where human interference is essential to ensure that users' rights are not abused on the grounds of incorrect HS charges, we must rely on metrics such as high recall and precision of HS category that we can guarantee to not overwhelm the human effort of checking redundant content. However, in such applications as HS reporting and handling, where human intervention is required to ensure that users' rights are not violated by false HS accusations, we should focus on metrics like high recall and $F_1$ score of the HS category, which ensure that human personnel checking redundant content are not overburdened.

\begin{table}[ht]
\centering
\resizebox{\textwidth}{!}{%
\begin{tabular}{| c | c | c | c | c | c | c | c | c |}
\hline
{\textbf{}} & {\textbf{$F_1$ Score}} & {\textbf{$F_1$ Hate}} & {\textbf{Accuracy}} & {\textbf{Precision}} & {\textbf{Sensitivity}} & {\textbf{Recall}} & {\textbf{Recall Hate}} & {\textbf{Specificity}} \\ \hline
{\textbf{MultinomialNB}} & {63.78} & {59.14} & {64.73} & {64.06} & {58.82} & {63.96} & {59.45} & {69.2} \\ \hline
{\textbf{BernoulliNB}} & {47.78} & {44.52} & {48.3} & {48.23} & {47.81} & {48.16} & {41.65} & {48.51} \\ \hline
{\textbf{Logistic Regression}} & {66.5} & {64.35} & {66.94} & {66.94} & {68.78} & {67.07} & {60.46} & {65.36} \\ \hline
{\textbf{SVM}} & {66.07} & {63.77} & {66.43} & {66.47} & {68.08} & {66.7} & {59.96} & {65.32} \\ \hline
{\textbf{Random Forests}} & {64.41} & {60.07} & {65.04} & {64.69} & {60.61} & {64.68} & {59.54} & {68.75} \\ \hline
{\textbf{Gradient Boosting}} & {63.55} & {59.21} & {64.33} & {64.34} & {59.67} & {64.2} & {58.76} & {68.73} \\ \hline
{\textbf{CNN+Attention+FT+GV}} & {75.76} & {71.76} & {76.56} & {76.86} & {68.64} & {75.66} & {75.18} & {82.68} \\ \hline
{\textbf{LSTM+FT+GV}} & {75.24} & {72.24} & {75.95} & {76.57} & {72.11} & {75.53} & {72.36} & {78.95} \\ \hline
{\textbf{FF+LSTM+CNN+FT+GV}} & {75.49} & {72.08} & {76.15} & {76.29} & {70.88} & {75.52} & {73.28} & {80.16} \\ \hline
{\textbf{BiLSTM+FT+GV}} & {77.84} & {75.40} & {78.16} & {78.05} & {77.15} & {78.04} & {73.73} & {78.94} \\ \hline
{\textbf{BERT}} & {79.60} & {77.13} & {79.96} & {79.89} & \textbf{{77.87}} & {79.73} & {76.4} & {81.59} \\ \hline
{\textbf{DistilBERT}} & \textbf{{79.92}} & \textbf{{77.16}} & \textbf{{80.36}} & \textbf{{80.28}} & {76.47} & \textbf{{79.91}} & \textbf{{77.87}} & \textbf{{83.36}} \\ \hline
\end{tabular}%
}
\caption{Performance of selected models on binary HS classification}
\label{tab:results1}
\end{table}

The handling of textual data is a thoroughly researched task and has a dedicated category, NLP, which stands for natural language processing. We used common and widely accepted techniques to process them, as described previously. In Table~\ref{tab:results1}, we are showcasing the results of the selected evaluation processes per each classifier. The best performance per metric is highlighted in bold format. The NNs seem to outperform the conventional ML techniques. It is worth mentioning that Bayesian learners had the lowest performance in terms of almost every metric, while tree-ensembles achieved similar performance between them, but lower compared to the SVMs and LR.

Between the examined NNs, those who achieved the highest performance using \emph{emb} were the architectures using BiLSTMs. BiLSTMs + FT + GV achieved the highest recall on hate category, as well as high accuracy. Finally, BERT and DistilBERT outperformed every other model in any metric, using fine-tuning on the data, with DistilBERT performing slightly better than BERT, validating its superior performance on similar tasks~\cite{DBLP:conf/fire/RanasingheZH19}.

\subsection{Multi-Label Classification}
Providing a dataset with multi-label information about HS, we are able to uncover new insights. HS is indeed an ML task that cannot be studied thoroughly just through the binary aspect. Indeed, it is a multi-dimensional task.

The algorithms handling MLL can be either problem transformation or adaptation techniques~\cite{tsoum}. MLkNN~\cite{zhang2007ml} and MLARAM~\cite{MLARAM}, as well as Binary Relevance (BR) and Classifier Chains (CC)~\cite{read2009classifier} with base learners like LR, SVMs and RF, are utilised. We used FT \emph{emb} for our NNs and designed models inspired by classic MLL systems, such as BR and CC. Specifically, NNBR is an NN containing BiLSTMs, an attention layer, two FFs and an output layer with 8 outputs in a BR fashion. NNCC is inspired by the CC technique, but during its output each label is given as input for the next label prediction.

\begin{table}[ht]
\centering
\resizebox{\textwidth}{!}{%
\begin{tabular}{|c|c|c|c|c|c|c|c|c|c|c|c|c|c|}
\hline
 & \textbf{\begin{tabular}[c]{@{}c@{}}$F_1$\\ Example\end{tabular}} & \textbf{\begin{tabular}[c]{@{}c@{}}$F_1$ \\ Macro\end{tabular}} & \textbf{\begin{tabular}[c]{@{}c@{}}$F_1$ \\ Micro\end{tabular}} & \textbf{\begin{tabular}[c]{@{}c@{}}P \\ Example\end{tabular}} & \textbf{\begin{tabular}[c]{@{}c@{}}P \\ Macro\end{tabular}} & \textbf{\begin{tabular}[c]{@{}c@{}}P \\ Micro\end{tabular}} & \textbf{\begin{tabular}[c]{@{}c@{}}R\\ Example\end{tabular}} & \textbf{\begin{tabular}[c]{@{}c@{}}R\\ Macro\end{tabular}} & \textbf{\begin{tabular}[c]{@{}c@{}}R\\ Micro\end{tabular}} & \textbf{\begin{tabular}[c]{@{}c@{}}AP\\ Macro\end{tabular}} & \textbf{\begin{tabular}[c]{@{}c@{}}AP\\ Micro\end{tabular}} & \textbf{\begin{tabular}[c]{@{}c@{}}Subset\\ Accuracy\end{tabular}} & \textbf{\begin{tabular}[c]{@{}c@{}}Hamming\\ Loss\end{tabular}} \\ \hline
\textbf{MLkNN} & 48.01 & 53.04 & 53.74 & 55.27 & 71.29 & 69.95 & 46.28 & 45.04 & 43.98 & 46.63 & 42.79 & 26.53 & 0.1566 \\ \hline
\textbf{MLARAM} & 18.47 & 6.06 & 18.71 & 21.44 & 3.78 & 21.44 & 17.69 & 16.25 & 18.27 & 20.79 & 21.55 & 7.15 & 0.2948 \\ \hline
\textbf{BR} & 48.59 & 52.49 & 56.76 & 57.69 & 79.74 & 79.37 & 45.30 & 42 & 44.37 & 47.66 & 47.04 & 26.28 & 0.1395 \\ \hline
\textbf{CC} & 56.51 & 59.24 & 58.23 & 62.49 & 69.08 & 63.44 & 56.54 & 56.22 & 53.99 & 49.74 & 44.07 & 31.4 & 0.1606 \\ \hline
\textbf{NNBR} & \textbf{75.05} & \textbf{76.23} & \textbf{74.87} & \textbf{81.02} & \textbf{83.21} & 79.27 & \textbf{74.33} & \textbf{73.04} & \textbf{71.29} & \textbf{67.33} & \textbf{62.64} & \textbf{48.39} & \textbf{0.0993} \\ \hline
\textbf{NNCC} & 47.66 & 51.25 & 55.47 & 57.34 & 73.36 & \textbf{84.27} & 44.06 & 42.40 & 41.70 & 50.02 & 47.36 & 26.61 & 0.1378 \\ \hline
\end{tabular}%
}
\caption{Performance of selected models on MLL HS (P: Precision, R: Recall, AP: Average Precision)}
\label{tab:mllresults}
\end{table}

In the evaluation of MLL systems, a very common measure is the Hamming loss (symmetric difference between the ground truth labels and the predicted ones). Furthermore, subset accuracy (symmetric similarity), as well as precision, recall and $F_1$-score, are contained here (instance-based metrics). Moreover, some label-based metrics like $B$-macro and $B$-micro, where $B\in\{$$F_1$, Precision, Recall$\}$ were computed. We present our results in Table~\ref{tab:mllresults}. The superior performance of neural-based approaches compared to classical ML models is observed. Specifically, NNBR achieves the highest score in 12 out of 13 metrics.

\section{Dataset Experimentation}
After setting the baseline performance of ETHOS in multiple ML algorithms, in both binary and multi-label scope, this section aims at highlighting some interesting views and aspects of its usefulness over other learning tasks. First, we fulfil our experimental soundness by setting a fair comparison between a balanced subset and a random subset of ETHOS capturing useful insights under a 1-vs-1 evaluation stage. Secondly, we examine how the ETHOS dataset can generalise over separate HS datasets when it is applicable. Thus, we transfer its discriminative ability obtained by the proposed underlying representation through training proper ML models. These experiments have been conducted for two well-known datasets on binary (D1)~\cite{hateTweets} (2017) and multi-label (D2)~\cite{DBLP:conf/emnlp/OusidhoumLZSY19} (2019) level, as described briefly in Section~\ref{sec:relDat}, commenting the produced results regarding the aspects that we had initially posed and providing accurate explanations about any mismatches over this attempt.

\subsection{Balanced vs Random Comparison}

Initially, we are going to experiment with the proposed dataset using just a few variations in the binary level. More precisely, we create two versions of ETHOS, one of which collects 75\% of data at random (DRa), while the other collects 75\% of data preserving the class balance (DBa), from a pool of 87.5\%. The remainder of the data (DRe), which is 12.5\%, will be used as test data. Two SVM models are then trained on DRa and DBa using a TFIDF vectorizer and evaluated on the DRe. We are running this experiment 10 times, shuffling appropriately our data. In addition, the two SVM models are evaluated on the D1 dataset as well. Under this scenario, we are further investigating the learning capacity of the constructed ETHOS dataset comparing two different variants: a strictly balanced and a random one, while our evaluation protocol is consistent with maintaining the balancing property of the generated sub-optimal subsamples. The application of the trained learners into separate datasets may confirm also our assumptions about the efficacy of our strategy: the active selection of multi-label samples for constructing a balanced HS dataset.

The results are shown in Table~\ref{tab:balancedVSrandom}, verifying that the performance of the SVM on the test set is higher when the dataset maintains a balance between classes. However, in terms of accuracy, a higher score is obtained by random datasets. We cannot conclude for the $F_1$ weighted performance of DRa and DBa on D1, as the wide standard deviation of the DBa makes it difficult. This result comes of course with an explanation: a defining characteristic of D1 dataset concerns its imbalanced nature. This indicates that the SVM trained on random data is more biased towards the majority class. In order to investigate this, the weighted $F_1$ score per label is shown in Table~\ref{tab:balancedVSrandomWhy}. 

\begin{table}[ht]
\centering
\resizebox{0.66\textwidth}{!}{%
\begin{tabular}{c|c|c|c}
\cline{2-3}
 & DRe & D1 &  \\ \hline
\multicolumn{1}{|c|}{Train on DRa} & 63.15 $\pm$ 3.93 & \textbf{50.62} $\pm$ 1.10 & \multicolumn{1}{c|}{\multirow{2}{*}{Accuracy}} \\ \cline{1-3}
\multicolumn{1}{|c|}{Train on DBa} & \textbf{67.99} $\pm$ 2.17 & 43.61 $\pm$ 12.39 & \multicolumn{1}{c|}{} \\ \hline
\multicolumn{1}{|c|}{Train on DRa} & 64.19 $\pm$ 4.89 & 36.15 $\pm$ 1.05 & \multicolumn{1}{c|}{\multirow{2}{*}{$F_1$ weighted}} \\ \cline{1-3}
\multicolumn{1}{|c|}{Train on DBa} & \textbf{69.06} $\pm$ 2.29 & {37.21} $\pm$ 8.25 & \multicolumn{1}{c|}{} \\ \hline
\end{tabular}
}
\caption{Comparison of SVM performance (metric$\pm$std) trained on random and balanced subsets of ETHOS and tested on unknown data from the same source (DRe) and a different one (D1)}
\label{tab:balancedVSrandom}
\end{table}

\begin{table}[ht]
\centering
\resizebox{0.47\textwidth}{!}{%
\begin{tabular}{c|c|c}
\cline{2-2}
 & D1 &  \\ \hline
\multicolumn{1}{|c|}{Train on DRa} & \textbf{66.53} $\pm$ 1.01 & \multicolumn{1}{c|}{\multirow{2}{*}{\begin{tabular}[c]{@{}c@{}}$F_1$ \\ Non HS\end{tabular}}} \\ \cline{1-2}
\multicolumn{1}{|c|}{Train on DBa} & 54.48 $\pm$ 16.32 & \multicolumn{1}{c|}{} \\ \hline
\multicolumn{1}{|c|}{Train on DRa} & 5.77 $\pm$ 19.94 & \multicolumn{1}{c|}{\multirow{2}{*}{\begin{tabular}[c]{@{}c@{}}$F_1$ \\ HS\end{tabular}}} \\ \cline{1-2}
\multicolumn{1}{|c|}{Train on DBa} & \textbf{19.94} $\pm$ 3.34 & \multicolumn{1}{c|}{} \\ \hline
\end{tabular}
}
\caption{Performance of SVM (metric$\pm$std) on D1 per label}
\label{tab:balancedVSrandomWhy}
\end{table}

As we previously assumed, the SVM model trained on DRa has a bias towards the majority class (No Hate) obtaining a better score than the SVM model trained on DBa. However, this is not the case for the minority class, which seems to be best predicted by the SVM trained on the DBa. In tasks such as hate speech identification, it would be more valuable to more precisely identify comments of hate speech. Consequently, a balanced dataset despite its limited cardinality may play a crucial role in tackling this phenomenon, verifying the assets of the proposed protocol.

\subsection{Generalising on binary level}

In an attempt to prove that a small but carefully collected dataset is of higher quality and more useful than larger datasets collected under unknown conditions, we will compare ETHOS to D1, a dataset 24 times larger. In this cross-validation experiment, we train an SVM model (with default parameters) on the ETHOS dataset and predict the D1 dataset, and vice versa. We have also computed the performance of SVMs on the D1 through nested cross-validation, resulting in 66.18\% balanced accuracy, 68.77\% $F_1$ weighted score, 96.97\% $F_1$ on non HS tweets and 42.09\% on HS tweets, revealing thus its optimal performance which also did not manage to get improved regarding the predictiveness of HS instances.

The results of each cross validation training are shown in Table~\ref{tab:SVMonETHOS} and Table~\ref{tab:SVMonD1}. It is visible that both SVMs perform equally in both metrics. It could be expected that the SVM trained on D1, a larger dataset, would perform better than a smaller dataset, but the more sophisticated manner of collecting and annotating data in the case of ETHOS overcomes its limited cardinality offering similar predictive ability with a quite larger collection of instances.

\begin{figure}[ht]
\centering
\minipage{0.46\textwidth}
\centering
\resizebox{1\textwidth}{!}{%
\begin{tabular}{c|c|c|} \cline{2-3}
 & ETHOS & D1\\ \hline 
\multicolumn{1}{|c|}{Balanced Accuracy} & 58.03 & 54.03 \\ \hline 
\multicolumn{1}{|c|}{$F_1$ weighted} & 56.41 & 87.32 \\ \hline 
\multicolumn{1}{|c|}{$F_1$ Non HS} & 74.03 & 91.88 \\ \hline 
\multicolumn{1}{|c|}{$F_1$ HS} & 33.21 & 12.85 \\ \hline 
\end{tabular}}
\captionof{table}{SVM model trained on ETHOS and predicting D1}
\label{tab:SVMonETHOS}
\endminipage\hfill
\minipage{0.46\textwidth}
\centering
\resizebox{1\textwidth}{!}{%
\begin{tabular}{c|c|c|} \cline{2-3}
 & D1 & ETHOS \\\hline 
\multicolumn{1}{|c|}{Balanced Accuracy} &  50.90 & 53.33 \\\hline 
\multicolumn{1}{|c|}{$F_1$ weighted} & 42.67 & 92.31 \\\hline 
\multicolumn{1}{|c|}{$F_1$ Non HS} & 72.66 & 97.10 \\\hline 
\multicolumn{1}{|c|}{$F_1$ HS} & 3.53 & 12.38 \\ \hline 
\end{tabular}}
\captionof{table}{SVM model trained on D1 and predicting ETHOS}
\label{tab:SVMonD1}
\endminipage
\end{figure}

It is peculiar that the two models do not predict the other's hate speech instances. Digging into that further, we can see that there are few problematic instances in D1. For example, the following sentence: \textit{``realdonaldtrump he looks like reg memphis tn trash we got them everywhere''} does not contain hate speech content, rather than offensive. Moreover, the distribution of the hate instances to hate categories in D1 is non-uniform, favouring three categories: race (dark-skinned people), sexual orientation (homosexual people) and gender (women). The aforementioned conclusion was the product of applying the ETHOS Multi-labelled dataset, predicting 326 - race, 257 - sexuality and 230 - gender instances out of the 1430 hate speech tweets, as well as the product of a simple word frequency calculation, suggesting that there are 378 - race (words: `n****r', `n***a', `n****h'), 417 - sexuality (words: `f****t', `f*g', `g*y', `q***r') and 352 - gender (words: `b***h', `c**t', `h*e') instances.

Finally, it would be interesting to investigate the overall performance of an SVM model trained on a combination dataset of those two. After a 10-fold cross validation training the combined dataset achieved 55.27\% balanced accuracy, 90.88\% $F_1$ weighted score, 96.48\% $F_1$ on Non Hate Speech and 18.84\% $F_1$ on Hate Speech. The overall performance of the model increased, implying that combining datasets with different dynamics can lead to better models. To this aspect, one of the posed ambitions of our work seems to be satisfied, since its integration with the D1 dataset leads to improved learning behaviour. 

\subsection{Generalising on multi-label level}

The dataset of ETHOS has two variants, a binary and a multi-labelled dataset. After experimenting with the binary version of it, we use the D2 dataset in this Section to show the usefulness of ETHOS. D2 is a multilingual and multi-aspect hate speech dataset containing information for tweets such as hostility type, directness, target attribute and category, as well as annotator's sentiment. However, there is no one-to-one mapping between these attributes and the attributes of ETHOS. For example, the type of hostility defines the sentiment of a tweet as abusive, hateful, offensive, disrespectful, fearful and normal. We assign instances described as abusive, hateful or fearful as violent, while others are described as non-violent. The mapping of the hostility directness to the ETHOS directed\_vs\_generalized label is straightforward. Finally, the mapping between the hate categories and the target attributes is almost the same, while the `race' category is absent. However, by extracting information from the target group attribute, we assign tweets to the `race' category when the target group is either `African descent' or `Asian'.

\begin{table}[ht]
\resizebox{1\textwidth}{!}{
\begin{tabular}{c|c|c|c|c|c|c|c|c|}
\cline{2-9}
\textit{} & Violence & \begin{tabular}[c]{@{}c@{}}Directed vs\\ Generalized\end{tabular} & Gender  & Race    & \begin{tabular}[c]{@{}c@{}}National\\ Origin\end{tabular}  & Disability & Religion & \begin{tabular}[c]{@{}c@{}}Sexual\\ Orientation\end{tabular} \\ \hline
\multicolumn{1}{|c|}{Accuracy} & 50.86  & 55.28  & 70.34 & 75.97 & 67.88 & 69.64 & 71.65  & 89.83 \\ \hline
\multicolumn{1}{|c|}{$F_1$ weighted}       & 59.48  & 55.39  & 87.71 & 92.78 & 68.97 & 83.81 & 97.65  & 94.21 \\ \hline
\multicolumn{1}{|c|}{$F_1$ (negative)}     & 72.50  & 59.36  & 92.94 & 94.61 & 74.89 & 91.06 & 98.51  & 96.50 \\ \hline
\multicolumn{1}{|c|}{$F_1$ (positive)}     & 29.09  & 19.98  & 46.59 & 24.06 & 61.23 & 53.44 & 27.31  & 71.29 \\ \hline
\end{tabular}}
\captionof{table}{The performance of the model trained on ETHOS predicting the labels of D2. \emph{$F_1$ (negative): The label is not appearing in the instance, $F_1$ (positive): The label is appearing in the instance}}
\label{tab:ETHOSonD2}
\end{table}

Training a neural network with BiLSTM layers using ETHOS multi-labelled dataset, we are predicting the labels of D2. In Table~\ref{tab:ETHOSonD2}, the performance of the model on the D2 dataset per label is showcased. The model achieved to predict perfectly the `sexual\_orientation' label, decently the `disability', `national\_origin' and `gender' labels, but poorly the `directed\_vs\_generalized', `violence', `race' and `religion' labels. Specifically, on the `religion' label the model can identify if a tweet does not contain hate speech towards religion by 97.82\%, but its performance is downgraded on the opposite case, achieving 27.31\%. About the `violence' label, the model fails to predict when a sentence incites violence with 29.09\%. The worst predicted label by the model is the `directed\_vs\_generalized'. This means that the model cannot generalise well when a tweet is targeting a specific individual.

As it regards the `race', due to the lack of information in the D2 about this label, it was expected to counter such a low performance. To be more convenient with this aspect, we depict some of the instances which had as groups `Asian' or `African descent', and our model did not categorise as race the following four:

\begin{quote}
    ``well my parents like carikla ching chong guy in your college'' \\
    ``yay kelas ching chong today''\\ 
    ``okay ching chong''\\
    ``remember it was some ching chong hoe on here that was flexin on him years ago found out they was fuckin smh''
\end{quote}

It seems the BiLSTM model has not encountered such examples. Indeed, ETHOS dataset does not contain any example with the phrase `ching chong'. However, we should investigate the reversed situation as well, namely, the instances that did not have the race label, but the BiLSTM model assigned it erroneously. This misclassification occurred to 35 instances, while 26 of them contain hate speech targeting `race'. We present here the most representative of them:

\begin{quote}
    ``see the type of n****r you are hmph''\\ 
    ``die n****r'' and 20 similar\\ 
    ``now yes this politically motivated terrorist is white and leftist'' and 3 similar
\end{quote}

Such issues are quite possible to occur because of mismatching between the separate collections of data. Enrichment of the source dataset, in our case the ETHOS, by a careful selection of instances that describe such cases could help our attempt. Therefore, the adoption of metric learning mechanisms may help us alleviate the hubness phenomenon which put obstacles on recovering distinct classes~\cite{DBLP:conf/cvpr/KimKCK20}.

\section{Discussion}

The provision of a new well-designed dataset to the public on a specific subject is always considered a significant contribution~\cite{DBLP:conf/cikm/HoangVN18,DBLP:conf/cikm/SunAJHS19}. In this sense, our HS dataset, called ETHOS, collected from social media platforms, could be reused by the ML and AI communities. Alleviating redundant information through balancing the proposed dataset between fine-grained classes through a fine-tuned learner and an Active Learning scheme benefited us both from the aspect of less human-laborious effort and, of course, by scoring good learning rates despite the limited cardinality of our collected instances. Redundancy reduction has been shown to be quite beneficial for a variety of learning tasks. More specifically, the proposed protocol offers us a balanced dataset with a rich quality of included instances for both binary and multi-label HS problems. At the same time, our experimental procedure revealed that a proper balance has been achieved between the discriminative ability of the learners, both traditional and neural networks, and the computational resources consumed.

The issue of imbalanced data collection has also affected the performance of similar works, where the need for proper manipulation is clearly stated~\cite{DBLP:conf/lrec/HaagsmaBN20,DBLP:conf/eacl/NghiemBA21}. The solution of proactive learning has been applied in the latter approach, trying to match the expertise of each human annotator with the most appropriate unlabelled instances. Based on this, the negative effect of harmful annotations can be seriously avoided. This asset should be carefully explored and adopted by our side before enlargement of the current dataset takes place or new data collection attempts get started. We must emphasise once more that, despite the relatively small size of the ETHOS dataset, the human resources invested in adequate labelling cannot be overlooked (2 consecutive months of daily querying of the targeted databases, human annotation in 2 stages, input by a crowdsourcing process). Thus, besides the need for high-quality annotators, mining informative instances that retain the ability to discriminate between hate speech examples, both in binary and multi-label classification tasks, is of high importance. The conducted experiments verify our assumptions following our straightforward protocol, since the learning performance of various models is satisfactory, especially these based on embeddings. Simultaneously, a proof-of-concept of how to exploit the ETHOS dataset's learning capacity was provided, serving as a seed dataset for generalising to similar hate speech detection datasets.

Some promising directions of our work are mentioned here, trying to take advantage of its assets and the baselines that were posed. The main issue, the shortage of collected data, is a fact that depends on the limitations that occur during exploiting crowdsourcing platforms (e.g. restricted budget, users' traffic) and the further costs that are induced by the human-intensive stage of actively selecting instances that keep a balanced profile of the target dataset on a daily basis. Investigating the related literature, we have mined some clever ideas that tackle this limitation. We record here the case where an annotation process has been designed using a game-based approach, motivating the human oracles to contribute to assigning sentiment labels to a variety of Twitter instances, surpassing the monetisation incentive~\cite{DBLP:journals/puc/FuriniM18}. Further enrichment of this dataset could also be carried out, integrating either multilingual resources for capturing even more hate speech occurrences, or applying data augmentation techniques~\cite{DBLP:conf/sac/ShimLLV20}. From the perspective of the ML models that we used, pre-processing stages -- such as feature selection mechanisms~\cite{DBLP:journals/inffus/TommaselG18} or methods for creation of semantic features~\cite{DBLP:journals/csl/SkrljMKLP21} -- which are established in the realm of short-text input data, could improve the obtained results and retain interpretability properties in specific cases.

In addition, the ETHOS can be combined with various similar HS datasets -- as we stated here initially with two different data collections -- for evaluation reasons. The development of hybrid weakly supervised HS detection models, merging semi-supervised and active learning strategies under common frameworks, alleviating human intervention based on decisions over the gathered unlabelled instances that come solely from the side of a robust learner~\cite{DBLP:conf/iisa/KarlosKAFK19,DBLP:journals/mlc/YuFXQ19}, constitutes another very promising ambition. Online HS detection and prevention tools, such as Hatebusters among others, are highly favoured by such approaches. The impact of such detection tools could have been very beneficial in terms of enforcing social awareness and addressing effective ethical issues~\cite{DBLP:conf/sai/AlharthiR20,DBLP:conf/ijcai/DinakarPL15}.

Finally, the fact of examining ETHOS under the spectrum of multi-labelled nature appears favouring to reviewers on social media platforms, facilitating informative suggestions for HS comments regarding the level of violence, the target of comments and the categories of HS that are present. However, this is not a multi-purpose HS detection dataset, as the mined comments are based on social media. This means that the corpus contains relatively small sentences. Thus, models trained on this dataset may fail to detect HS in documents on a larger scale without segmentation. On the other hand, the general structure of the proposed protocol could be applied to a variety of learning tasks, especially on large databases, towards better predictions and less intensive annotation~\cite{DBLP:journals/biomedsem/DrameMD16}. Last but not least, examination of alternative query sampling strategies that support inherent MLL could have proven quite beneficial regarding both the reduction of human effort and the enrichment of attempts like the proposed one~\cite{DBLP:journals/jcst/KumarG20}.

\bibliographystyle{spmpsci}

\end{document}